%% file: main.tex
\documentclass[sigplan,screen]{acmart}
\AtBeginDocument{%
  \providecommand\BibTeX{{%
    \normalfont B\kern-0.5em{\scshape i\kern-0.25em b}\kern-0.8em\TeX}}}
    
\usepackage{bm}
\usepackage{mathtools}
\usepackage{algorithm2e}
\SetKw{KwBy}{by}
\newcommand{\argmin}[1]{\underset{#1}{\operatorname{arg}\,\operatorname{min}}\;}

\copyrightyear{2022}
\acmYear{2022}
\setcopyright{rightsretained}
\acmConference[EuroMLSys'22]{2nd European Workshop on Machine Learning and Systems }{April 5--8, 2022}{RENNES, France}
\acmBooktitle{2nd European Workshop on Machine Learning and Systems (EuroMLSys'22), April 5--8, 2022, RENNES, France}\acmDOI{10.1145/3517207.3526980}
\acmISBN{978-1-4503-9254-9/22/04}

\begin{document}

\title{Data Selection for Efficient Model Update in Federated Learning}

\author{Hongrui Shi}
\email{hshi21@sheffield.ac.uk}
\affiliation{%
  \institution{University of Sheffield}
  \country{}
}

\author{Valentin Radu}
\email{valentin.radu@sheffield.ac.uk}
\affiliation{%
  \institution{University of Sheffield}
  \country{}
}

\renewcommand{\shortauthors}{Hongrui Shi and Valentin Radu}

\begin{abstract}
The Federated Learning (FL) workflow of training a centralized model with distributed data is growing in popularity. 
However, until recently, this was the realm of contributing clients with similar computing capability. The fast expanding IoT space and data being generated and processed at the edge are encouraging more effort into expanding federated learning to include heterogeneous systems. Previous approaches distribute light-weight models to clients are rely on knowledge transfer to distil the characteristic of local data in partitioned updates. However, their additional knowledge exchange transmitted through the network degrades the communication efficiency of FL. 
We propose to reduce the size of knowledge exchanged in these FL setups by clustering and selecting only the most representative bits of information from the clients. The partitioned global update adopted in our work splits the global deep neural network into a lower part for generic feature extraction and an upper part that is more sensitive to this selected client knowledge. Our experiments show that only 1.6\% of the initially exchanged data can effectively transfer the characteristic of the client data to the global model in our FL approach, using split networks. These preliminary results evolve our understanding of federated learning by demonstrating efficient training using strategically selected training samples.

\end{abstract}




\maketitle

\input{introduction}

\input{related_work}

\input{methods}
\input{evaluation}

\input{conclusions_future_work}

\bibliographystyle{ACM-Reference-Format}
\bibliography{sample-base}


\end{document}

%% file: introduction.tex
\section{Introduction}\label{sec:introduction}

Federated Learning (FL) is seen as the best solution for training machine learning models on private data of many devices. The widespread adoption of smartphones and their capability of generating essential training data have contributed to the emergence of this privacy-preserving learning method~\cite{mcmahan2017communication}. 
FL is used in many commercial applications, from training the Google Keyboard for next work prediction~\cite{kairouz2019advances} to on-mobile image classification and more~\cite{bonawitz2019towards}.

In the classic FL setting, a global model is copied from a server to many clients to be used for their local inference. A subset of clients are chosen to contribute to the update of the global model by sending their local training results on the local data to the server. The server aggregates all the local training results to update the global model at the end of a FL round. Subsequently, the updated global model is distributed to clients to start a new round.

This training process works well when most clients have similar computing capability. However, devices with a lot of local data or reduced computing resources will fall behind the other clients and stall the training round. To compensate for this system heterogeneity, recent research efforts have proposed solutions based on model compression~\cite{abdelmoniem2021towards} and student models trained with attention transfer~\cite{FLwithAT}. Their underlying idea is to reduce the computation load during the local updates by training a light-weight model. In~\cite{FLwithAT} we previously showed the effectiveness of using attention transfer and metadata in the form of activation maps from the Convolution Neural Network (CNN) to perform the model update in a heterogeneous FL environment. However, it fails in a critical point -- if a client harbors a large number of local samples, the amount of activation maps that needs to be sent to the server is critically large for network transfer. Consequently, this adds a significant overhead to the communication of our FL setting. 

To this end, we aim to reduce the size of activation maps needed for updating the global model, without deteriorating its generalization capability. Our proposed solution is to identify the most informative samples for the knowledge transfer. We do this by clustering and then select a smaller set of activation maps to be used for the final stage of the global model update at each round. 


We alter the federated learning setting by introducing a split CNN paradigm. The global CNN model is split into two parts, a lower part with layers extracting generic features, and an upper part that is more sensitive to the activation maps. For the lower part we keep a similar training process to the federated average method, aggregating local updates into the lower part of the global model. On the other hand, the upper part of the CNN model is trained using a reduced set of activation maps created by the lower part of the distributed global model on the local data. The adoption of activation maps for updating the upper part of the global model is build on the solution conceived in our previous work~\cite{FLwithAT} introduced for the challenge of heterogeneous devices. 


Our split training paradigm preserves the privacy policy of FL by sending a small fraction of activation maps to the server, and not the actual raw local data. This approach has been previously explored~\cite{osia2020hybrid} from a privacy perspective, and proving the fallibility of raw data reconstruction from feature maps is outside the scope of this paper. 

Essentially, we make the following contributions:

\begin{itemize}
    \item We propose a new FL setting that splits the global neural network model into two parts, a lower part for generic feature extraction trained with the federated average method, and an upper part updated through activation maps. We explore different levels for splitting the model, driven by empirical observations. 
    \item To improve the communication efficiency, we introduce a clustering approach to reduce the size of activation maps used for global update, without greatly deteriorating the performance of our global model. This work advances FL in heterogeneous systems, such as our previous system~\cite{FLwithAT}, because the amount of computation performed by less powerful devices can be significantly reduced using less local data. 
    
    \item By selecting representative samples from the clusters, our evaluation indicates that as little as 1.6\% of activation maps can update the global model effectively with a negligible drop in test accuracy on the Cifar-10 image classification~\cite{Krizhevsky2009}. However, There is still scope for improving these results and we indicate the route for this in future work presented in Section~\ref{sec:conclusions}.
\end{itemize}

%% file: related_work.tex

\section{Related Work}\label{sec:related_work}

\paragraph{\textbf{Federated Learning (FL)}}
FL aims to train a global model in massively distributed networks~\cite{mcmahan2017communication} at scale \cite{bonawitz2019towards}, over heterogeneous data of many sources~\cite{li2018federated}, and with different training approaches~\cite{FLwithAT,smith2017federated,he2020group}. For the system heterogeneity challenge of training using devices of various computing capability, some previous solutions abandon the single client model paradigm and replace this with compressed client models to run efficiently on the client by matching the model size with device computing capability~\cite{FLwithAT,abdelmoniem2021towards}.


\paragraph{\textbf{System Heterogeneity.}}

The differences in hardware characteristics (processor, frequency, memory, etc.) across clients and the amount of local data for training produce large variations in the number of local updates performed
by each client~\cite{wang2020tackling}. 
In FL, a deadline is imposed by the server on the time in order to allow the clients to finish their local updates. 
Those clients who manage only a few local updates or none at all in the given time are called \textit{stragglers} and often their work is discarded.
FedNova~\cite{wang2020tackling} aggregates even the updates from straggler by factoring in the number of local updates performed by each client for weighting their contribution to the update of the global model.

The other problem with system heterogeneity is that it causes objective inconsistency. The federated optimisation convergence is built on the assumption that clients perform a similar amount of updates from their local data~\cite{wang2018cooperative,stich2018local}. 
The best approach for assuring convergence is to allow all clients to finish their round. But waiting for the slow clients can significantly increase the training time~\cite{wang2020tackling}.

Some most recent works allow less powerful clients to finish their local updates in time by making the client models adaptive to local hardware. ~\cite{abdelmoniem2021towards} uses model compression technique to reduce the client models and ~\cite{FLwithAT} allows the client models to have different sizes using attention transfer and metadata training to aggregate local knowledge globally. 

\paragraph{\textbf{Non-IID data.}} 
FL is expected to learn from non-IID data across devices that generate vastly different distribution and volume of data. 
FedProx~\cite{li2018federated}, VRLSGD~\cite{liang2019variance} and SCAFFOLD~\cite{karimireddy2020scaffold} have been designed to handle non-IID local data. But these methods either result in slower convergence or require additional communication and memory.

Our previous work ~\cite{FLwithAT} shows that the metadata created by the client models can update the global model effectively in an extreme non-IID settings for local data. However, the metadata sent to the server will introduce additional overhead to the communication.


\paragraph{\textbf{Training paradigms.}} Different training paradigms have been explored for FL to address system and data heterogeneity. Formulating the training as a multi-task learning~\cite{smith2017federated} results in larger global models that can accumulate the multiple perspectives of client models trained on non-IID data as virtual tasks. Knowledge transfer is used by Liang et al.~\cite{liang2020think} to train split networks. Knowledge transfer is performed at the network separation point, running the first part of the model on the client and the final part on the server. Unlike~\cite{liang2020think}, we see benefit in having the entire global model on the server. This full model can be shared and used by clients that have no local data to train a portion for their model on.

To facilitate the knowledge transfer between the server and clients in FL,~\cite{FLwithAT} proposes to use attention transfer to train smaller models for each client from a large global model.



\paragraph{\textbf{Clustering}}

In the field of knowledge transfer, knowledge selection is proposed for seeking to select the most relevant knowledge for distillation. In~\cite{li2021knowledge}, an entropy-based threshold is introduced to decide whether a piece of knowledge is confident enough to be selected in a mutual knowledge distillation paradigm. 

Besides knowledge selection, another line of work is sample selection, where clustering techniques are used. Clustering is commonly used for data analysis as an unsupervised machine learning technique to group data points on the basis of their properties~\cite{Cluster}. Data points grouped in the same cluster are similar, whereas data points from different clusters have discriminating patterns. 

A popular clustering algorithm is K-means clustering. It clusters data points by minimizing the within-cluster squared Euclidean distances with Expectation-Maximum (EM) algorithm. In practice, K-means enjoys fast convergence by having a linear complexity. 
On the other hand, K-means suffers from two major problems. First, the number of clusters must be chosen manually, which is less intuitive in many cases. 
Secondly, the initialization of K-means may lead to the algorithm converging on different local minimums, making the results inconsistent. 

However, the computation efficiency of K-means makes it a good choice to cluster the representations in resource-constrained systems. 

\paragraph{\textbf{Meta Learning and Few-shot Learning}}

Meta Learning or Learning to Learn (LTL)~\cite{hochreiter2001learning} has emerged as a popular technique for model adaption. It relies on meta-knowledge extracted as generic information across tasks to generalise for a new task. Recently, Model Agnostic Meta Learning (MAML)~\citep{finn2017model} has stemmed from Meta Learning to learn a global model that can be used as an initialization for further learning of a good model adaptive to a new task by only a few local gradient steps \citep{kairouz2019advances}. Both \cite{khodak2019adaptive} and \cite{jiang2019improving} explore the relationship between MAML and FL in the purpose to address the heterogeneity problem, particularly the non-IID data. The global update in FL is compared to meta-training an initial model in MAML. The local updates performed by each client can be considered as the meta-testing to adapt the initial model to a individual local task.

To adapt the meta-learned initial model to a new task, Meta-Learning is typically followed by Few-Shot learning \citep{fei2006one}, which aims to use as less task-specify samples as possible. 
We see Few-Shot learning as another potential solution to update the global model with metadata of reduced size. This approach has not yet explored sufficiently in previous work as far as we know, so we are encouraged to apply this technique to our future work.

%% file: methods.tex

\begin{table}[t]
\caption{The list of notations used in this work.}\vspace{-0.5cm}
\begin{center}
\begin{tabular}{ll}
\toprule
    Notations & Description \\
    \midrule
    $t$                 & The $t$th global training round \\
    $M$                 & Total number of clients\\
    $D_{M_k}$           & Metadata from client $k$\\
    $D_{M}$             & The union of metadata from all clients\\
    $A^{[j]}_k$         & Activation maps from client $k$ at level $j$\\
    $M_{G}$             & The global model\\
    $M_{C_k}$           & The $k$th client model \\
    $M_{COM}$           & The composed model\\
    $\bm{W}_{G}$        & Weights of the global model \\
    $\bm{W}_{G}^{u}$    & Weights of upper part of the global model\\
    $\bm{W}_{G}^{l}$    & Weights of lower part of the global model\\
    $\bm{W}_{C_k}$      & Weights of the client model $k$\\
    $\bm{W}_S^{u}$      & Weights of upper part of the updated global model\\
    \bottomrule
\end{tabular}
\label{tab:notations}
\end{center}
\end{table}

\begin{figure}[t]
    \centering
    \includegraphics[width=0.9\linewidth]{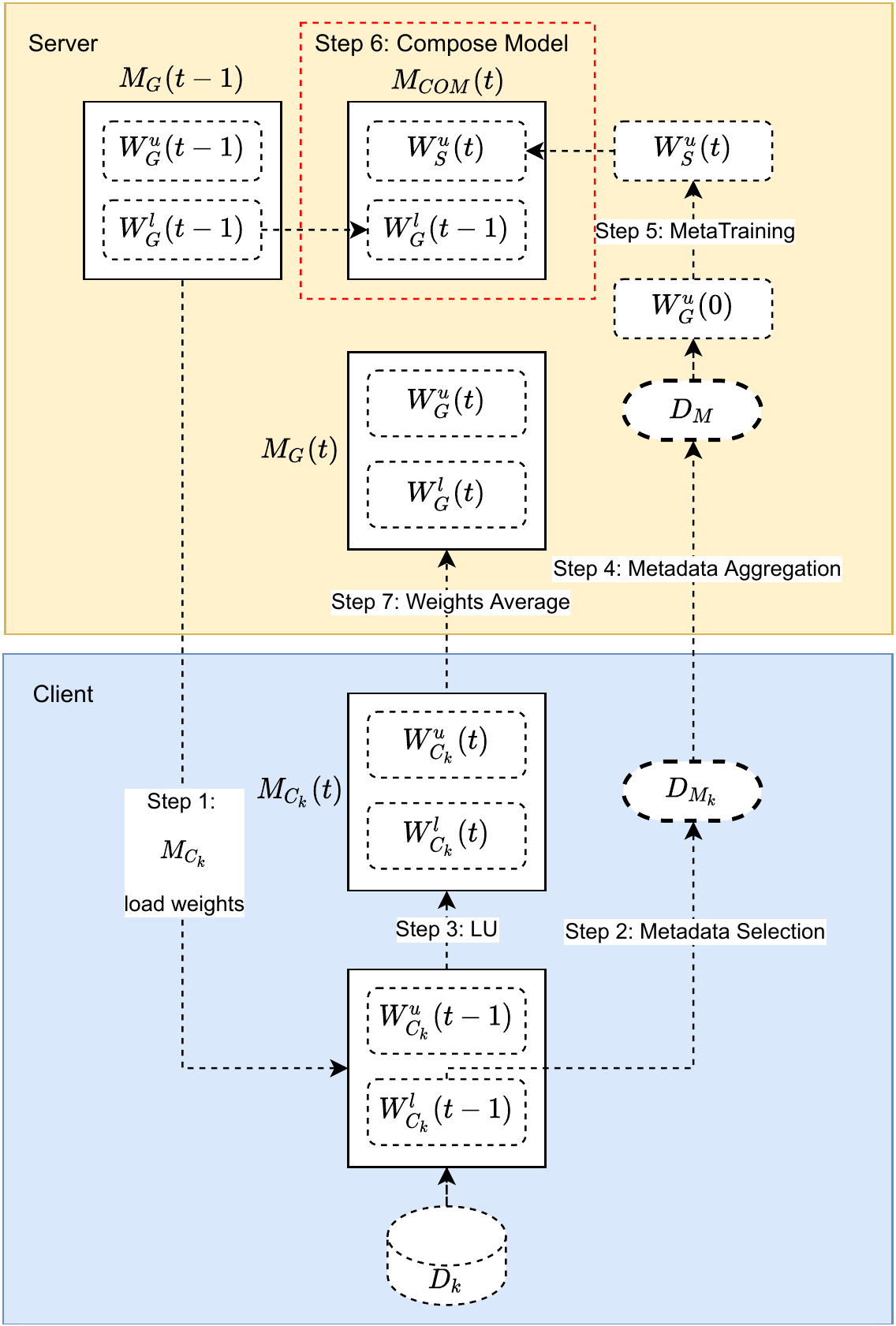}
    \caption{Block representation of the training performed between the server and clients.The model is split into two parts for efficient distributed training. The lower part of the network is trained using a federate learning approach. The upper part of the network is trained using activation maps (metadata) generated by clients using the local samples. A small fraction of activation maps can be identified as most relevant for training. We do this using clustering.}
    \label{fig:server_at}
\end{figure}

\RestyleAlgo{ruled}
\SetKwComment{Comment}{/* }{ */}
\begin{algorithm}
\caption{Split Training with Metadata Selection}\label{alg:two}
\textbf{Initialization} Initialize $M_{G}$ with weights $W_{G}(0)$\;
\For{each round $t=1, 2, ...$}{
\textbf{On the client side:}

    \For{each client $k$}{
    Client model $M_{C_k}$ loads weights $W_{G}(t-1)$\;
    $D_{M_k}(t) \leftarrow$ Extract\&Select $\left(D_{k}, W_{G}^{l}(t-1)\right)$\;
    $W_{C_k}(t) \leftarrow$ LocalUpdate $\left(D_{k}, W_{G}(t-1)\right)$\;
    }
\textbf{On the server side:}

Aggregate all $D_{M_k}(t)$ into $D_{M}(t)$\;
$W_{S}^{u}(t) \leftarrow$ MetaTraining $\left(D_{M}(t), W_{G}^{u}(0)\right)$\;

$M_{COM}(t) \leftarrow$ ModelCompose $\left(W_{G}^{l}(t-1), W_{S}^{u}(t)\right)$\;
Test $M_{COM}(t)$ on test dataset\;
$W_{G}(t) \leftarrow$ WeightAverage $\left(W_{C_k}(t), k=1, \cdots, m\right)$\;
}
return $W_{COM}$
\label{alg: Split Training}
\end{algorithm}

\section{Split Learning with Sample Selections}\label{methodology}
The global CNN model in our proposed method is trained on both the server side and the client side. The lower part of the model is trained in the standard federated average approach, with locally updated models being aggregated on the server side for updating the lower part. The upper part of the model is trained entirely on the server side, using the activation maps produced and uploaded by the clients. The list of the notations used in this section are presented in Table~\ref{tab:notations}.

\subsection{Activation Map Selection}\label{sec:selection}

\begin{figure}[t]
    \centering
    \includegraphics[width=0.4\linewidth]{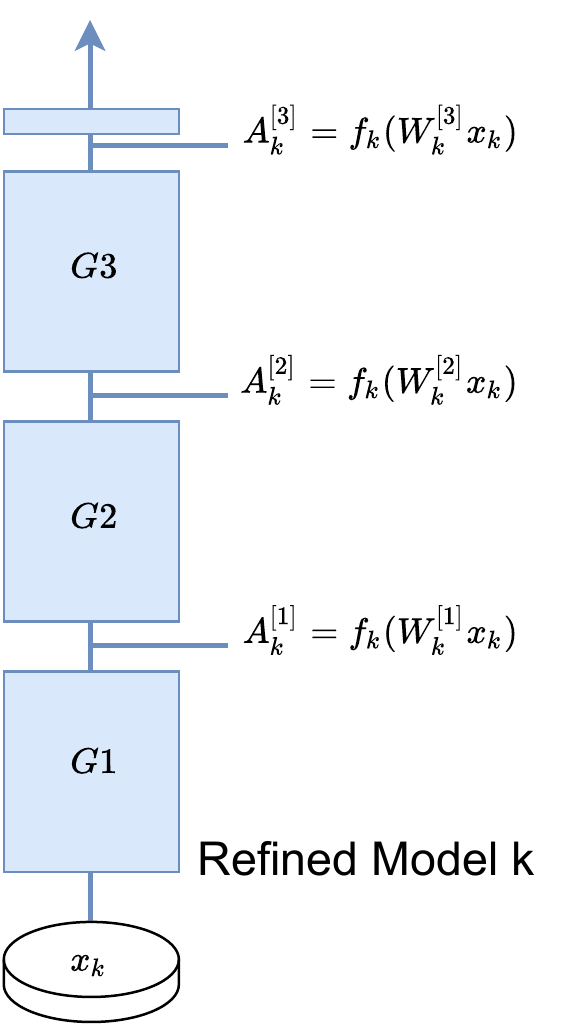}
    \caption{The collection of activation maps (metadata) from client model k (global model refined in the previous round). The level in the model from where to generate the activation maps is pre-determined and fix throughout the training.}
    \label{fig:metadata_extraction}
\end{figure}

At the start of each round $t$, the global model produced in the previous round $t-1$ is distributed to the client side as $M_{C_k}$. There, the activation maps (Equation (\ref{Eqa. metadata})) are created by the local data $D_{k}$ on client $k$, at a predefined level $j$ in the network that splits $M_{C_k}$ into a lower part and a upper part. 

\begin{equation}
    A^{[j]}_k=f_{C_k} (\bm{W}^{[j]}_{C_k}, \: x_k)
    \label{Eqa. metadata}
\end{equation}

\noindent where $f_{C_k}$ is the function represented by the client model $k$, $\bm{W}^{[j]}_{C_k}$ indicates the weights of the lower part of $M_{C_k}$ divided by level $j$, $(x_k, y_k)$ is the input-target pair of the local dataset $D_{k}$. Figure \ref{fig:metadata_extraction} illustrates the activation maps created by a CNN model at 3 different levels $j \in \{G1, G2, G3\}$ on the local data. Our level selection is similar to the one proposed in ~\cite{Zagoruyko2017}.

After the activation maps are created, we select the activation maps that are most representative alongside with their labels to form the metadata from client $k$:\\ $D_{M_k} = Select\left( A^{[j]}_k, y_k\right)$, sending them to the server for updating the upper part of the global model. The global model will then assimilate the local knowledge from the client models and local data encoded in this metadata.

To identify the most representative activation maps we propose to use Principal Component Analysis (PCA) for information compression and K-means for clustering. PCA is first used to reduce the dimension of the activation maps. Over the dimension-reduced activation maps, K-means is clusters these representations belonging to the same class label.

Our insight of using PCA+K-means is that activation maps with similar characteristics can be projected into the same cluster. Within each cluster, we choose the sample that is closest to the cluster centre with respect to the Euclidean distance as the most representative sample for its group. This selected sample is assumed to include the most common characteristics of all other samples in their own cluster. Finally, the metadata of client $k$,  $D_{M_k}$, is the union of the activation maps that are identified as the most representative samples in the clustering stage.

The number of expected clusters is a predetermined hyperparameter. If we set a small number of clusters, the size of metadata can be significantly reduced. Most importantly, the selected activation maps are assumed to be capable of updating the global model effectively without a significant loss of global generalization because they contain the most common local information.

\subsection{Local Update on the Client Side}

The global model delivered to the $k$-th client is tuned on the local data $D_{k}$ over a few local epochs. From these local epochs, the client optimiser adjusts the weights $\bm{W}_G(t-1)$ locally such that it minimises the loss function $\ell_k$ over the local data:

\begin{equation}
\begin{split}
    &\argmin{\bm{W}_{G}(t-1)} F \left(\bm{W}_{G}(t-1), \; D_{k}\right) \\
    &= \frac{1}{N_k}\sum_{i=1}^{N_k} \ell_k\left(f_{G}(\bm{W}_{G}(t-1),\; x_k^{(i)}),\; y_k^{(i)}\right)
    \end{split}
\end{equation}

\noindent where $f_{G}$ is the function represented by the global model, $N_k$ is the size of $D_{k}$, and $(x_k^{(i)}, y_k^{(i)})$ is the $i$-th instance of input-target pair of the local dataset. After the local updates, $\bm{W}_G(t-1)$ is uniquely updated in the $k$-th client to $\bm{W}_{C_k}(t)$. 

\subsection{Composing the Global Model on the Server Side}\label{sec:composed model}

On the server side, the level for partitioning the lower part and upper part of the global model is predetermined by $j$, using the same value for creating the activation maps introduced in Section~\ref{sec:selection}. Level $j$ is kept unchanged across global model and clients so that $D_{m}$ is able to match the input dimensions of the upper part of the global model. The server updates the upper part and lower part separately by using metadata training and federated average.

The upper part of the global model is updated to $\bm{W}_S^{u}(t)$ using the metadata $D_{m} = \bigcup\limits_{k=1}^{m}D_{M_k}$ collected from the clients at round $t$. To strengthen knowledge transfer for each round in the upper layers, we keep the same initialization $\bm{W}_G^{u}(0)$ for the upper part of the global model. Using a good number of metadata training epochs ensures the global model effectively update its upper part. The optimization objective of the metadata training is formulated as follows.

\begin{equation}
\underset{\bm{W}_S^{u}(t)}{\operatorname{argmin}} F \left(\bm{W}_S^{u}(t), \;D_{m}\right) =  \underset{\bm{W}_S^{u}(t)}{\operatorname{argmin}} \sum_{i=1}^{M} \ell_{G}\left(f_G(\bm{W}_S^{u}(t), \; A^{[j]}_i), \;y_i\right)
\label{Eqa. metadata training}
\end{equation}

\noindent where $f_G$ is the function represented by the global model, $M$ is the total number of clients, $(A^{[j]}_i, y_i)$ is the metadata from the $i$-th client in $D_{m}$.

To update the lower part of the global model, we adopt the conventional federated average approach. In addition to the metadata training, the server also gets an updated global model $M_{G}(t)$ by averaging the locally updated model, $\bm{W}_{C_k}(t)$, from all clients as follows:

\begin{equation}
    \bm{W}_{G}(t) = \frac{1}{m} \sum_{k=1}^{m} \bm{W}_{C_k}(t)
\end{equation}

Consequently, $M_{G}(t)$ contains the updated weights of the lower part. Finally, the server composes a new global model $M_{COM}(t)$, with its upper part from $\bm{W}_S^{u}(t)$ and lower part from $\bm{W}_G^{l}(t-1)$. It should be noted that we use the lower part from the global model from the previous round. This is because the metadata used to update the upper part to $\bm{W}_S^{u}(t)$ is essentially created by the global model from the previous round $M_{G}(t-1)$.  

At the end of each global round, the composed global model, $M_{COM}(t)$, is tested on the test dataset to evaluate its performance. The global model from federated average, $M_{G}(t)$, is distributed to the clients for local updates in the next round. 

Our algorithm is summarized in Algorithm \ref{alg: Split Training} and Figure \ref{fig:server_at} presents a block representation of the algorithm.

%% file: evaluation.tex
\section{Evaluation}\label{sec:evaluation}
This section presents the setup for validating our proposed solution, results and analysis of our experiments. 

\subsection{Experiment Setup}

Our experiments are conducted on the CIFAR-10 dataset~\cite{Krizhevsky2009} for the image classification task. CIFAR-10 has a total number of 60,000 images of size $32\times 32$ pixels, with 10 classes and 6,000 images for each class. The training set and test set in CIFAR-10 contain 50,000 images and 10,000 images respectively. Our results indicate the accuracy on the test set throughout the evaluate for the global model.

The Wide Residual Networks (WRN)~\cite{zagoruyko2016wide} is used as the global CNN model. The global model $M_G$ is built on a WRN with a depth of 40 and width of 1 (WRN-40-1). The layers of WRN-40-1 are hierarchically organized into 3 groups as discussed in Section \ref{sec:selection}. The dimension of activation maps extracted at different level $j$ of the WRN-40-1 is shown in Table \ref{tab:dimension of metadata}. 

\begin{table}[]
\begin{center}
\caption{The dimension of the activation maps at different levels in the global CNN model (WRN-40-1). For an image input, the higher the level of the model, the lower resolutions of the activation map.}
\begin{tabular}{@{}cc|c@{}}
\toprule
 Global Model  &  Level $j$ & Dimension of $A^{[j]}$  \\ \midrule
 WRN-40-1 & \begin{tabular}[c]{@{}c@{}}  $G1$\\$G2$\\$G3$\end{tabular} &
  \begin{tabular}[c]{@{}c@{}}  $16 \times 32 \times 32$\\$32 \times 16 \times 16$\\$64 \times 8 \times 8$\end{tabular}  \\ \midrule
\end{tabular}

\label{tab:dimension of metadata}
\end{center}
\end{table}


We borrow the experiment settings from our previous work to approach the heterogeneous system challenge~\cite{FLwithAT}. We set the number of clients to 20 and assume that all the clients are able to complete their local updates and metadata selection in time before finishing the round. To make a fair comparison, the baselines also assumes a full client participation. To simulate real-world applications, we exercise an extreme non-IID local data setting used in~\cite{Liang2020}, where each client hosts only 2500 images randomly selected from two random classes of CIFAR-10.

\subsection{Results and Analysis}\label{sec:Results and Analysis}

We fix the level in the network from where we collect the activation maps. This is first set to level $j = G1$, after the first group of convolutional layers.

The first experiments assess the effectiveness of training the model with selected metadata. We use the SGD optimizer with a learning rate of 0.1 and a batch size of 50 for model update. L2 regularization with a weight of 0.0005 is applied to metadata training. The training epochs are set to 1 and 100 respectively for the local updates and metadata training. We train the global model for 100 FL rounds. Regarding the hyperparameters of our proposed clustering method introduced in Section \ref{sec:selection}, the number of PCA components is set to 200, reducing the dimension of activation maps from $16 \times 32 \times 32$ to just 200 features. Dimension-reduced activation maps from the same (image recognition) class are clustered into 20 groups by K-means. As such, each CIFAR-10 class in $D_{k}$ will contribute 20 representative samples towards the metadata $D_{M_k}$. 

To evaluate the effect of using metadata selection, we use another baseline produced by using the entire set of activation maps from the whole training set on the composed model $D_{COM}$ (full metadata). This baseline collects all the activation maps produced by the clients to form $D_{M}$. Table~\ref{tab:withorwithoutselection} compares our proposed metadata selection to the baseline of using the entire set of metadata (activation maps). When the upper part of $M_{COM}$ is updated on the entire activation maps without selection (50000 in total), $D_{COM}$ achieves a higher test accuracy of 70.03\%. In contrast, the test accuracy of $M_{COM}$ is just 48.47\% due to $D_{M}$ being reduced to containing just 800 activation maps selected by our proposed clustering approach. A drop in accuracy is understandable due to reduced training data. We aim to narrow this performance gap by evaluating different levels in the network for extracting activation maps. 

\begin{table}[]
\begin{center}
\caption{Comparison of the performance of $D_{COM}$ with metadata selection and without metadata selection (full activation maps to train on) on the CIFAR-10 image classification task. In metadata selection, each client selects 40 activation maps to form the metadata, uniformly distributed across classes. This is 1.6\% of total activation maps available over the entire training set. The option without selection has no clusters.}
\begin{tabular}{@{}c|cccc@{}}
\toprule
 Composed Model  & \begin{tabular}[c]{@{}c@{}}Metadata\\Selection \end{tabular} & \begin{tabular}[c]{@{}c@{}}Cluster\\ Number \end{tabular} &Test Acc. \\ \midrule
 WRN-40-1  &
  \begin{tabular}[c]{@{}c@{}}without\\clustering\end{tabular} &
  \begin{tabular}[c]{@{}c@{}}  ---------\\20\end{tabular} &
  \begin{tabular}[c]{@{}c@{}} 70.03\%\\48.47\%\end{tabular} \\ \midrule
\end{tabular}

\label{tab:withorwithoutselection}
\end{center}
\end{table}

\subsection{Metadata from Higher Levels}\label{sec:overfitting_problem}

\begin{table}[t]
\begin{center}
\caption{Comparison among centralized training, Federated Average (FedAvg) and our FL approach with selected metadata. All use WRN-40-1 as the global model. The centralized version trains the model using the entire training data on a single machine, thus achieving the best performance. The FedAvg forms another strong baseline by assuming full client participation in each round. In our approach, we show the performance of the composed model trained on full activation maps and selected activation maps with on our clustering based selection (PCA+K-means). Our FL approach using split learning and metadata training outperforms the FedAvg baseline if all activation maps from either level $G1$ or $G2$ are used. With respect to using the selected metadata, we show that using selected metadata from level $G3$ produces a global model with a minimal performance drop. Missing values indicate not applicable settings. 
}
\begin{tabular}{@{}cc|c@{}}
\toprule
Method & \begin{tabular}[c]{@{}c@{}}Metadata\\Selection \end{tabular} & \begin{tabular}[c]{@{}c@{}}Test Acc.\\ \midrule \begin{tabular}{@{}cccc@{}}Global&$j=G1$&$j=G2$ &$j=G3$\end{tabular} \end{tabular} \\ \midrule
Centralized & -----------   & \begin{tabular}{@{}cccc@{}}90.79\%  & ----------- & -----------& ----------- \end{tabular}  \\ \midrule
FedAvg & -----------   & \begin{tabular}{@{}cccc@{}}69.35\%  & ----------- & -----------& ----------- \end{tabular}  \\ \midrule
\begin{tabular}[c]{@{}c@{}}FL+metadata\\ 
FL+metadata\end{tabular} & \begin{tabular}[c]{@{}c@{}}without\\ 
clustering\end{tabular} &
  \begin{tabular}[c]{@{}c@{}} \begin{tabular}{@{}cccc@{}}----------- & 70.03\% & 73.47\%& 68.55\%\end{tabular}\\ 
  \begin{tabular}{@{}cccc@{}}----------- & 48.47\%&60.11\%& 64.20\% \end{tabular}\end{tabular} \\ \midrule
\end{tabular}

\label{tab:different group level}
\end{center}
\end{table}

The performance of the composed model $M_{COM}$ whose upper part is trained on the selected activation maps from $G1$ of the client models degrades significantly compared to the baseline. Our intuition for this behaviour is that the selected metadata from a lower level is insufficient to train the upper part of the WRN-40-1 model, which has more than half a million parameters. 

To justify this hypothesis, we further reduce the number of learnable parameters in the upper part by predefining a higher $j$ level. Table \ref{tab:different group level} shows our experiment for using metadata from all three levels. In addition to the baseline without using metadata selection described in Section \ref{sec:Results and Analysis}, we further include a baseline where the global model is trained on the entire CIFAR-10 training set in a centralized manner and a FedAvg ~\cite{mcmahan2017communication} baseline assuming full client participation.

The result depicts a narrow performance gap when a large $j$ is chosen. With a fraction of activation maps (1.6\% of all activation maps) from level $G3$ for updating the global model, the performance of the model decreases by less than 5\%. In contrast, using selected metadata from a lower level leads to a significant performance deterioration. However, we notice that using a larger $j$ makes our approach resemble the standard FedAvg because the lower part of the global model, which is averaged in our approach, has a large size. This observation comes also from the experimental result which show that our method without metadata selection achieves 68.55\%. This performance is very close to the performance of the FedAvg baseline of 69.35\%

On the other hand, if metadata from level $G1$ or $G2$ is fully used, our FL approach outperforms the FedAvg baseline on the test set by over 4\%. Compared with averaging the entire global model, this result clearly demonstrates the advantage of using metadata to update the upper part of the split global model.

%% file: conclusions_future_work.tex
\section{Conclusions and Future Work}\label{sec:conclusions}

We propose a new split training approach for federated learning, with the aim to reduce the size of metadata (activation maps) required from the client side to train a global model. This approach splits the global CNN model into a lower part that is trained using the federated learning standard approach, and an upper part of the network that is trained on the server side using a fraction of metadata selected from the client side. We select the metadata by determining the most representative samples through clustering. Our experiments show that with selecting just 1.6\% of the activation maps, the composed global model maintains a good performance, decreasing just 4.35\% in image classification accuracy on the CIFAR-10 test set. Our approach improves the communication efficiency of our previous work to benefit federated learning in heterogeneous systems due to reducing the local data needed for updates. 

There is still scope for improving our results. In future, we will explore two main research directions:

\begin{itemize}
    \item replacing the distance-based clustering selection of representative activation maps with stronger similarity metrics.
    \item exploring more effective methods to train large models (holding many parameters) with very few training samples, inspired from few-shot learning.
\end{itemize}


